\documentclass[journal]{IEEEtran}  

\usepackage{graphics} 
\usepackage{epsfig} 
\usepackage{mathptmx} 
\usepackage{times} 
\usepackage{amsmath} 
\usepackage{amssymb}  
\usepackage{multirow}
\usepackage{array} 
\usepackage{booktabs}
\usepackage{caption}
\usepackage{subcaption}
\usepackage{graphicx}
\usepackage{xcolor}
\usepackage{threeparttable}
\usepackage[hidelinks]{hyperref}

\begin{document}
\title{\LARGE 
Transformer-Based Fault-Tolerant Control for Fixed-Wing UAVs Using Knowledge Distillation and In-Context Adaptation
}

\author{Francisco Giral$^{1}$, Ignacio Gómez$^{1}$, Ricardo Vinuesa$^{2}$, Soledad Le Clainche$^{1}$
\thanks{$^{1}$Francisco Giral, Ignacio Gomez and Soledad Le-Clainche are with the Applied Mathematics Department at the ETSIAE-School of Aeronautics, Universidad Politécnica de Madrid, 28040, Madrid, Spain. (e-mail: fa.giral@alumnos.upm.es, ignacio.gomez@upm.es, soledad.lechainche@upm.es)}%
\thanks{$^{2}$Ricardo Vinuesa is with FLOW, Engineering Mechanics, KTH Royal Institute of Technology, Stockholm 100 44, Sweden. (e-mail: rvinuesa@mech.kth.se)}
}



\maketitle

\begin{abstract}

This study presents a transformer-based approach for fault-tolerant control in fixed-wing Unmanned Aerial Vehicles (UAVs), designed to adapt in real time to dynamic changes caused by structural damage or actuator failures. Unlike traditional Flight Control Systems (FCSs) that rely on classical control theory and struggle under severe alterations in dynamics, our method directly maps outer-loop reference values—altitude, heading, and airspeed—into control commands using the in-context learning and attention mechanisms of transformers, thus bypassing inner-loop controllers and fault-detection layers. Employing a teacher-student knowledge distillation framework, the proposed approach trains a student agent with partial observations by transferring knowledge from a privileged expert agent with full observability, enabling robust performance across diverse failure scenarios. Experimental results demonstrate that our transformer-based controller outperforms industry-standard FCS and state-of-the-art reinforcement learning (RL) methods, maintaining high tracking accuracy and stability in nominal conditions and extreme failure cases, highlighting its potential for enhancing UAV operational safety and reliability.
Video: \href{https://youtu.be/ATW3LZFRqc0}{https://youtu.be/ATW3LZFRqc0}

\end{abstract}


\section{Introduction}

\IEEEPARstart{I}{n} recent years, Unmanned Aerial Vehicles (UAVs) have been widely used to perform various applications in complex and critical scenarios, such as search and rescue or autonomous medical transportation. The operational safety and reliability of these aerial robots have become major concerns due to the potential implications of system failures.

\begin{figure*}[htbp]
  \centering
  \begin{subfigure}{0.47\textwidth}
    \centering
    \includegraphics[width=\linewidth]{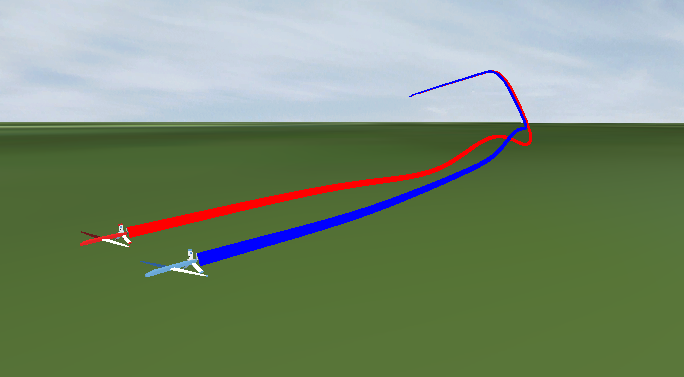}
    \caption{}
    \label{fig:traj_1}
  \end{subfigure}
  \begin{subfigure}{0.47\textwidth}
    \centering
    \includegraphics[width=\linewidth]{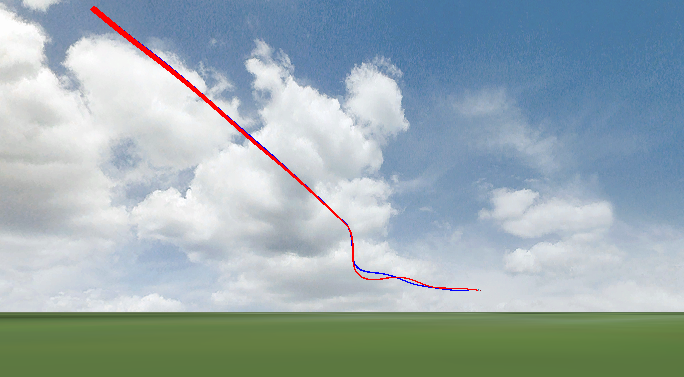}
    \caption{}
    \label{fig:traj_2}
  \end{subfigure}
  \vspace{0.5cm} 
  \begin{subfigure}{0.47\textwidth}
    \centering
    \includegraphics[width=\linewidth]{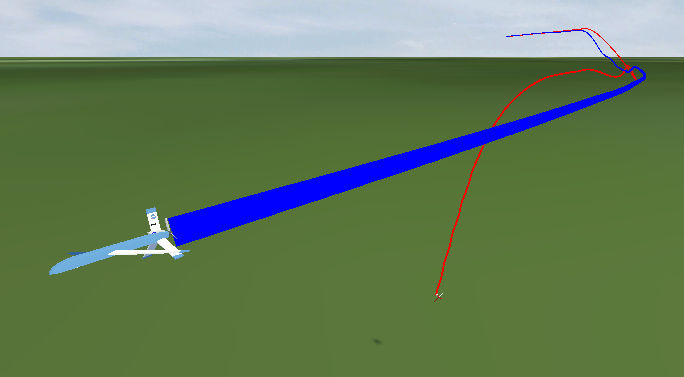}
    \caption{}
    \label{fig:traj_3}
  \end{subfigure}
  \begin{subfigure}{0.47\textwidth}
    \centering
    \includegraphics[width=\linewidth]{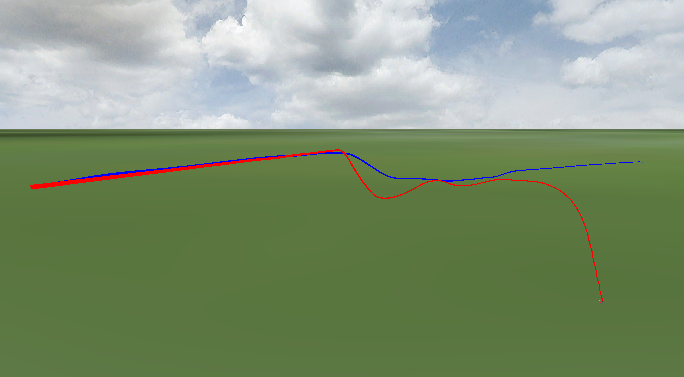}
    \caption{}
    \label{fig:traj_4}
  \end{subfigure}
  \caption{Trajectory comparison between the proposed transformer-based controller (blue) and an industry-standard FCS (red). Figures (a) and (b) illustrate nominal scenario tracking, while (c) and (d) demonstrate the controllers’ responses to semi-wing damage, with the FCS losing control and the proposed method stabilizing the UAV}
  \label{fig:unity_trajectories}
\end{figure*}

In contrast to robotics fields like manipulation and humanoid locomotion—which rely on advanced control methods to manage complex joint movements—UAV flight control systems (FCSs) in industry typically use classical control techniques for low-level control. Although modern approaches such as Model Predictive Control (MPC) offer significant advantages for high-level tasks like trajectory planning and collision avoidance \cite{celestini2022trajectory, lindqvist2020nonlinear}, they require precise system models, extensive uncertainty handling, and high computational resources, which often makes them impractical for low-level UAV control. Instead, the simplicity, reliability, and efficiency of classical control techniques—such as cascade Proportional-Integral-Derivative (PID) control systems—have made them the preferred choice for UAV attitude control, and they are implemented in widely deployed autopilots like PX4 and Ardupilot.

However, complex environments and demanding tasks can cause structural damage to the UAV, altering its aerodynamic characteristics and dynamics. Fixed-wing UAVs, in particular, exhibit highly complex, nonlinear dynamics, which can be significantly disrupted if the structure is compromised. Although current FCSs are robust, they struggle to maintain performance when the vehicle dynamics deviate from the original design specifications, sometimes leading to control divergence and catastrophic failure.

Fault-tolerant flight control has become a focal point for safety-critical UAV operations. Typically, fault-tolerant methods rely on fault detection and diagnosis techniques, which identify faults and then adjust controller parameters to account for the new dynamics \cite{ergoccmen2020active, mueller2014stability, miao2020dynamics, o2024learning}. This approach is complex, requiring real-time fault identification and parameter adjustment in a highly nonlinear dynamic environment. Other studies have explored novel approaches to nonlinear fault-tolerant control without explicit fault identification. These include approaches based on fuzzy logic systems \cite{yao2019fault}, sliding mode control \cite{ezzara2024sliding}, and robust control techniques \cite{cao2024performance}.

Reinforcement Learning (RL) has introduced alternative solutions to the problem \cite{dally2022soft, gavra2024evolutionary, liu2024reinforcement, tonti2024navigation}. With its ability to handle high-dimensional, nonlinear dynamics \cite{Vignon_2023}, RL holds promise for system fault management. However, RL algorithms are typically designed for Markov Decision Process (MDP) formulations, while fault-tolerant control—with sudden, unobserved changes in dynamics—must be framed as a Partially Observable Markov Decision Process (POMDP), making RL algorithms learning more difficult and possibly leading them to suboptimal performance \cite{ni2021recurrent}. Although recent works have shown possible solutions to this problem \cite{ni2021recurrent, hafner2023mastering, parisotto2020stabilizing}, these add considerable complexity to the algorithms. Most research focuses on actuator faults in multicopter UAVs \cite{mueller2014stability,liu2024reinforcement} or on fixed-wing UAV fault tolerance at the inner attitude control loop level, avoiding altitude, heading, and airspeed tracking \cite{dally2022soft, gavra2024evolutionary}.

Although learning-based control methods like Reinforcement Learning (RL) and Imitation Learning (IL) have great potential, ensuring stability with these techniques is more complex than with traditional control theory algorithms due to their intrinsic non-linearity and the lack of a model of the system. Several studies have attempted to integrate stability guarantees into RL and IL-based algorithms \cite{han2020actor, yin2021imitation}. Other works, based on Lyapunov theory, focus on learning a candidate Lyapunov function--commonly referred to as a Neural Lyapunov function \cite{liu2024tool, zhou2022neural, boffi2021learning, dawson2022safecontrollearnedcertificates}. In Ref. \cite{deka2023supervised}, instead of simply penalizing violations of the Lyapunov condition without labeled data for direct regression, the authors utilize numerical tools based on Koopman Operator theory to generate explicit ground truth data for training.

In this work, we propose a novel transformer-based fault-tolerant control method to directly map reference points in the outer loop of attitude UAV control—altitude ($h$), heading ($\Psi$), and airspeed ($V_T$)—into control-surface and throttle commands, enabling full UAV control without the need for inner control loops, which complicate the system. Our method employs the attention mechanisms and in-context learning capabilities of transformer models to adapt control actions to dynamic changes during inference, thereby eliminating the need for fault detection or identification and parameter adjustments in the Flight Control System (FCS). Our transformer-based controller uses a context window of past UAV states to autonomously detect and adapt to dynamic changes, similar to the ideas described in Ref. \cite{chen2024trakdis}. Moreover, we incorporate Lyapunov-based concepts into the offline RL training process to encourage stable model behavior. Fig. \ref{fig:unity_trajectories} shows a comparison between our proposed transformer-based controller (blue trajectory) and an industry-standard FCS (red trajectory). In Fig. \ref{fig:traj_1} and Fig. \ref{fig:traj_2}, both systems track the commanded references in nominal conditions. Figs. \ref{fig:traj_3} and \ref{fig:traj_4} illustrate the response when the UAV experiences semi-wing damage, showing the FCS losing control and causing a crash, while our method stabilizes the UAV and follows the reference values.

The contributions of this work are as follows:

\begin{itemize}
    \item We present a novel learning framework based on teacher-student knowledge distillation for learning fault-tolerant policies in fixed-wing UAVs. In this framework, the teacher agent is trained using RL on privileged environment information to address partial observability limitations, then a student agent is trained without privileged information using the teacher's interactions with the environment.
    \item We design a transformer-based flight controller that utilizes in-context learning to adapt its behavior in real-time as UAV dynamics change due to failures. This controller employs a context of past states to determine actions, eliminating the need for fault detection methods. Moreover, Lyapunov-based concepts are included to encourage controller stability.
    \item We conduct a comparative study against state-of-the-art methods, addressing not only actuator faults but also significant structural damage scenarios.
\end{itemize}

\section{Related Works}

\subsection{Learning-Based Fault-Tolerant Control of Aerial Vehicles}

Due to the complex dynamics and limited controllability, the problem of fault-tolerant flight control remains challenging. Numerous attempts have been made to address this issue, with data-driven methods gaining popularity in recent years.

Many studies focus on multicopters, often addressing actuator faults \cite{o2024learning,liu2024reinforcement,dooraki2020reinforcement}. These works demonstrate the solvability of this problem, either through pre-trained fault-sensing networks \cite{o2024learning} or RL approaches \cite{dooraki2020reinforcement}, even when multiple motors fail.

For fixed-wing UAVs, while several studies have applied RL methods to flight-attitude control \cite{de2023deep,li2023basic}, relatively fewer have focused specifically on fault-tolerant control, which must address a broader range of potential faults beyond actuator failures. An early exploration into using RL for low-level fault-tolerant flight control is presented in Ref.~\cite{dally2022soft}, where the authors employ an actor-critic architecture within an inner-loop controller. This approach represents an important step towards RL-based resilience in flight control, with a focus on maintaining stability at lower levels of control. In Ref.~\cite{gavra2024evolutionary}, RL is combined with Genetic Algorithms (GA) to train a population of agents, offering a novel method that introduces added implementation complexity and highlights the potential of hybrid approaches.

\subsection{Transformers}

Transformers, known for their attention mechanism and advantages in sequence modeling, were initially applied in the field of Natural Language Processing (NLP) \cite{vaswani2017attention}. Owing to their high-quality global contextual learning and efficient parallel computation, the transformer architecture has emerged as a powerful alternative to traditional sequence prediction methods like Recurrent Neural Networks (RNNs) and Long Short-Term Memory Networks (LSTMs). From a robotics and control point of view, transformers outperform recurrent models in capturing long-range dependencies and rapidly adapting to dynamic system variations, as demonstrated in Ref. \cite{radosavovic2024real}. Later in this work, a performance comparison between these models is conducted.

With advances in transformers, their application has extended to RL \cite{chen2021decision, zheng2022online, janner2021offlinereinforcementlearningbig}. Researchers have employed transformer architectures in decision-making agents, proposing the Decision Transformer (DT) model, which frames the sequence modeling problem as an action prediction problem based on historical states and reward-to-go sequences \cite{chen2021decision}. Trained in a supervised fashion, DT seeks to minimize the error between predicted and ground-truth actions by using sequences of past states and reward-to-go. Recent work has shown that DTs can outperform many state-of-the-art offline RL algorithms \cite{chen2021decision}. In this work, a DT model is trained to serve as an attitude controller for a UAV in failure scenarios, outperforming other state-of-the-art methods.

\subsection{Knowledge Distillation}

Knowledge distillation is a transfer learning (TL) approach that aims to transfer knowledge from a teacher model to a student model \cite{zhu2023transfer}. The original concept of knowledge distillation (KD) involves transferring knowledge from a large, complex model (the teacher) to a smaller, simpler model (the student) \cite{hinton2015distilling}. This transfer is typically achieved by minimizing the Kullback-Leibler (KL) divergence between the teacher's and student’s outputs.

Recent work has combined TL with memory components to address partial observability. In Ref.~\cite{tirumala2024learning}, this combination is used to train a student policy to play soccer based on partial visual observations. In Ref.~\cite{caluwaerts2023barkour}, multiple teachers are distilled into a single multitask student. Authors in Refs.~\cite{chen2024trakdis} and \cite{radosavovic2024real} distill knowledge into a transformer-based student policy to manage partial observability in manipulation tasks and humanoid locomotion, respectively. In contrast, in this work, expert knowledge from the privileged agent is distilled into the DT student in a non-privileged manner, enabling it to adapt its behavior in real time to failures affecting UAV dynamics.

\section{Problem Statement}

In this work, we aim to address the problem of fault-tolerant control in fixed-wing UAVs, where, beyond merely avoiding a crash after a failure, the vehicle is still able to track the given reference values. At a high level of flight-attitude control, we consider the problem of tracking specified set-points in altitude ($h_{\rm ref}$), heading ($\Psi_{\rm ref}$), and airspeed ($V_{T_{\rm ref}}$).  

We consider faults that significantly alter the aerodynamic characteristics of the UAV, such as a damaged control surface or a damaged wing. These failures primarily affect the stability and control derivatives, leading to substantial changes in the vehicle's dynamics. Stability and control derivatives define the forces and moments acting on the UAV, commonly expressed as:

\begin{align}
    C_L &= C_{L_0} + C_{L_\alpha} \alpha + C_{L_q} \frac{qc}{2V} + C_{L_{\delta_e}} \delta_e \\
    C_D &= C_{D_0} + k C_L^2 + C_{D_{\delta_e}} \delta_e \\
    C_Y &= C_{Y_\beta} \beta + C_{Y_p} \frac{pb}{2V} + C_{Y_r} \frac{rb}{2V} + C_{Y_{\delta_a}} \delta_a + C_{Y_{\delta_r}} \delta_r \\
    C_l &= C_{l_0} + C_{l_\beta} \beta + C_{l_p} \frac{pb}{2V} + C_{l_r} \frac{rb}{2V} + C_{l_{\delta_a}} \delta_a + C_{l_{\delta_r}} \delta_r \\
    C_m &= C_{m_0} + C_{m_\alpha} \alpha + C_{m_q} \frac{qc}{2V} + C_{m_{\delta_e}} \delta_e \\
    C_n &= C_{n_0} + C_{n_\beta} \beta + C_{n_p} \frac{pb}{2V} + C_{n_r} \frac{rb}{2V} + C_{n_{\delta_a}} \delta_a + C_{n_{\delta_r}} \delta_r
\end{align}

\vspace{0.1cm}

These equations define the aerodynamic coefficients $C_L$, $C_D$, $C_Y$, $C_l$, $C_m$, and $C_n$, which represent the lift, drag, side-force, roll moment, pitch moment, and yaw moment coefficients, respectively \cite{stevens2015aircraft, low2025modeling}. Each equation is influenced by the UAV's dynamic variables (e.g., angle of attack $\alpha$, sideslip angle $\beta$, pitch rate $q$, roll rate $p$, yaw rate $r$) and control surface deflections (e.g., elevator $\delta_e$, aileron $\delta_a$, rudder $\delta_r$). In these expressions, $c$ refers to the mean aerodynamic chord, $b$ is the wingspan of the UAV, $k$ is a factor related to induced drag, and $V$ is the UAV's airspeed. The coefficients are functions of the stability and control derivatives, which are significantly affected by the UAV's aerodynamic configuration and can be substantially altered in cases of faults or damage.

The aerodynamic forces and moments derived from the stability and control derivatives dictate the time evolution of the state variables, typically represented as a nonlinear, highly coupled dynamical system, $\dot{x} = f(x, u)$.

Current state-of-the-art control algorithms are unable to adapt to the changes in vehicle dynamics once failure occurs. Therefore, faults and damages in the UAV significantly impact the performance of existing flight control laws, sometimes leading to catastrophic outcomes \cite{matsuki2018flight}.

To overcome these limitations, we propose a learning-based controller leveraging the power of a transformer model. This new model can use a history of the UAV's state variables to adapt its behavior in context, without updating its weights.

\section{Fault-Tolerant Flight Control Via Sequence Modeling}

We formulate the flight-control-under-failure conditions as a partially observable Markov decision process (POMDP). The POMDP is represented by the tuple $\langle A, S, O, P, R \rangle$, consisting of actions $a \in A$, states $s \in S$, observations $o \in O$, transition functions $p \in P$, and rewards $r \in R$. At each timestep $t$, the RL agent receives an observation $o_t$ and generates an action $a_t$. Then the environment is updated based on the transition function $p(s_{t+1} | s_t, a_t)$ and assigns a reward to the agent. The agent’s objective is to maximize cumulative discounted rewards $R_t = \sum_{t=k}^{H} \gamma^t r_t$, where $H$, $k$, and $\gamma$ are the horizon length, current timestep, and discount factor, respectively.

Common POMDP tasks involve challenges where observations only partially reflect the underlying state, where different states may appear identical, or where observations are affected by random noise. Key subareas within POMDP formulations in RL include Meta-RL, Robust RL, and Generalization in RL \cite{ni2021recurrent}. In these subareas, agents are designed to handle variations in dynamics and unseen test environments by learning adaptive behaviors. For instance, Meta-RL addresses tasks with episode-specific variations in dynamics, while Robust RL focuses on environments with adversarial perturbations, and Generalization in RL emphasizes resilience to out-of-distribution states \cite{ni2021recurrent}. In the context of UAV control under damage and failure, the problem aligns with these subareas, as it involves sudden, substantial shifts in dynamics that are not directly observable, requiring the control policy to adapt without explicit knowledge of the altered dynamics.

\subsection{Privileged Agent Training through Online Reinforcement Learning}

The primary goal of the RL agent is to minimize the error between the actual values and commanded set-points for the three variables that allow complete control of the UAV in space: altitude ($h$), heading ($\Psi$), and airspeed ($V_T$).

The agent takes actions to modify the UAV’s dynamics using control surfaces—aileron ($\delta_a$), elevator ($\delta_e$), and rudder ($\delta_r$)—and throttle ($\delta_T$). These actions are based on a set of observations, $o \in O$, consisting of tracking errors ($\epsilon_h$, $\epsilon_{\Psi}$, $\epsilon_{V_T}$), altitude ($h$), aerodynamic angles ($\alpha$, $\beta$), attitude angles ($\phi$, $\theta$), angular velocity ($p, q, r$), linear velocity ($u, v, w$), and linear accelerations ($n_x, n_y, n_z$). These observations define the UAV’s attitude.

The reward function is designed to track the reference values while maintaining smooth attitude adjustments and minimizing control effort. The total reward is calculated as a weighted sum of these components:
\vspace{-1mm}
\begin{equation}
    \text{r} = \lambda_1 \cdot r_{\text{h}} + \lambda_2 \cdot r_{\Psi} + \lambda_3 \cdot r_{V_T} + \lambda_4 \cdot r_{\text{att}} + \lambda_5 \cdot r_{\delta},
\end{equation}
where $r_{\text{h}}$, $r_{\Psi}$, and $r_{V_T}$ are tracking rewards, $r_{\text{att}}$ encourages a smooth attitude, and $r_{\delta}$ penalizes large control inputs. The weights $\lambda_1, \lambda_2, \lambda_3, \lambda_4, \lambda_5$ are set to $0.24, 0.2, 0.16, 0.2, 0.2$, respectively.

Tracking reward components take the form:

\begin{equation}
    r_{x} = -1 + \exp\left(-k \cdot \left|\frac{\epsilon_{x}}{\lambda_{\text{s}}}\right|\right),
\end{equation}
where $x$ represents $h$, $\Psi$, or $V_T$, and $k$ and $\lambda_{s}$ are shaping parameters. All reward weights and shaping parameters are initially chosen based on domain knowledge and are further tuned through trial and error to achieve the desired performance.

The attitude reward is calculated using angular rates ($p, q, r$) and roll angle ($\phi$) as:

\begin{equation} \label{expr:reward_att}
    r_{\text{att}} = -1 + \left(\hat{r_p} \cdot \hat{r_q} \cdot \hat{r_r} \cdot \hat{r_{\phi}}\right)^{1/4},
\end{equation}
while the control reward uses the deviation of the control commands with respect to the previous time step:

\begin{equation} \label{expr:reward_control}
    r_{\delta} = -1 + \left(\hat{r_{\Delta\delta_a}} \cdot \hat{r_{\Delta\delta_e}} \cdot \hat{r_{\Delta\delta_r}} \cdot \hat{r_{\Delta\delta_T}}\right)^{1/4} ,
\end{equation}

For expressions (\ref{expr:reward_att})–(\ref{expr:reward_control}), $\hat{r_x}$ takes the form:

\begin{equation}
    \hat{r_x} = \exp\left(-\left(\frac{x}{\lambda_s}\right)^2\right),
\end{equation}
where $x \in [p, q, r, \phi, \Delta\delta_a, \Delta\delta_e, \Delta\delta_r, \Delta\delta_T]$.

\vspace{0.2cm}

\begin{figure}[thpb]
  \centering
  \includegraphics[scale=0.17]{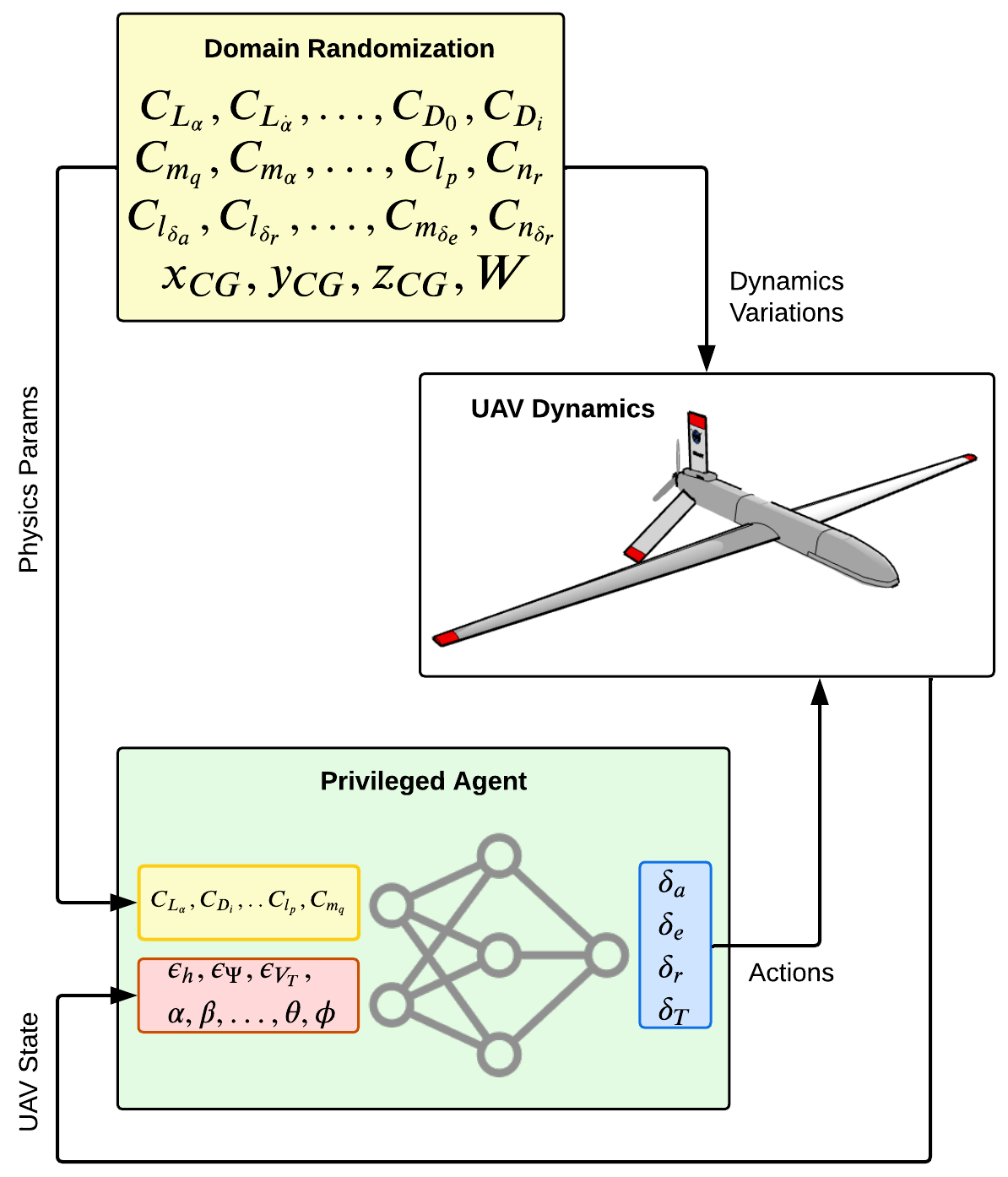}
    \caption{Online training process of the RL algorithm using Domain Randomization and privileged information to enhance adaptability and robustness under varying dynamics. The Domain Randomization (DR) block applies transformations to the physical parameters of the environment, modifying the dynamics. The privileged agent receives as input the concatenation of the current physics parameters and the UAV state obtained from the environment. Based on this information, the agent takes actions over the control surfaces ($\delta_a, \delta_e, \delta_r$) and throttle ($\delta_T$).}

  \label{fig:training_process}
  \vspace{-4mm}
\end{figure}

To address partial observability, a privileged RL agent is trained. By adding the physical parameters that characterize the environment to the agent's observations, the POMDP problem is transformed into a classical MDP, resulting in a more efficient and robust agent. As illustrated in Fig. \ref{fig:training_process}, Domain Randomization (DR) is applied to the environment’s physical parameters, altering the UAV's dynamics at the start and throughout each episode. By combining the agent’s partial observations ($o_t$) with the environment’s physical parameters at each timestep, forming the complete state ($s_t$), the agent learns to operate across varying UAV configurations and dynamics influenced by aerodynamic, center of gravity, and weight changes across episodes.

The modified physical parameters and their randomization ranges are shown in Table \ref{tab:domain_randomization}. Each range is selected based on known physical transformations caused by UAV failure or damage. During training, initial and reference flight conditions ($h$, $V_T$) are also randomized within the UAV’s flight envelope.

The privileged agent is trained using the DreamerV3 algorithm \cite{hafner2023mastering}, selected for its efficiency, robustness, and ability to generalize across multiple environments without extensive hyperparameter tuning. DreamerV3 is an online model-based RL algorithm that learns a world model of the environment from past trajectories, then refines a policy through imagination across the latent space learned in the world model encoder. \footnote{After publication, the code will be available at: \href{https://github.com/fgiral000/ft-decision-transformer}{https://github.com/fgiral000/ft-decision-transformer}}

\begin{table}[ht]
\caption{Domain Randomization Ranges}
\label{tab:domain_randomization}
\begin{center}
\begin{threeparttable}
\begin{tabular}{|c|c|}
\hline
\textbf{Parameter} & \textbf{Randomization range} \\ \hline \hline
$C_{L_{\alpha}}$, $C_{L_{\dot{q}}}$, $C_{L_{\dot{\alpha}}}$ & Scaling \(\mathcal{U}(0.6, 1.0)\) \\ \hline
$C_{D_{0}}$, $C_{D_{i}}$ & Scaling \(\mathcal{U}(1.0, 3.0)\) \\ \hline
$C_{D_{mach}}$ & Scaling \(\mathcal{U}(1.0, 2.0)\) \\ \hline
$C_{D_{\beta}}$ & Scaling \(\mathcal{U}(1.0, 3.0)\) \\ \hline
$C_{Y_{\beta}}$ & Scaling \(\mathcal{U}(0.5, 3.0)\) \\ \hline
$C_{Y_{\dot{p}}}$, $C_{l_{\beta}}$, $C_{l_{p}}$, $C_{l_{r}}$, $C_{n_{\alpha}}$, $C_{n_{\beta}}$, $C_{n_{\dot{p}}}$ & Scaling \(\mathcal{U}(0.5, 3.0)\) \\ \hline
$C_{Y_{\dot{r}}}$ & Scaling \(\mathcal{U}(0.5, 2.0)\) \\ \hline
$C_{m_{0}}$ & Additive \(\mathcal{U}(-0.02, 0.001)\) \\ \hline
$C_{m_{\alpha}}$ & Scaling \(\mathcal{U}(0.5, 1.5)\) \\ \hline
$C_{m_{q}}$ & Scaling \(\mathcal{U}(1.0, 3.0)\) \\ \hline
$C_{m_{\dot{\alpha}}}$ & Scaling \(\mathcal{U}(1.0, 1.5)\) \\ \hline
$C_{l_{0}}$, $C_{n_{0}}$ & Additive \(\mathcal{U}(-0.02, 0.009)\) \\ \hline
$C_{n_{r}}$ & Scaling \(\mathcal{U}(0.5, 1.0)\) \\ \hline
$C_{L_{\delta_e}}$, $C_{D_{\delta_e}}$, $C_{m_{\delta_e}}$, $C_{l_{\delta_a}}$, $C_{n_{\delta_a}}$ & Scaling \(\mathcal{U}(0.5, 1.0)\) \\ \hline
$C_{l_{\delta_r}}$, $C_{n_{\delta_r}}$, $C_{Y_{\delta_r}}$ & Scaling \(\mathcal{U}(0.0, 1.0)\) \\ \hline
$x_{CG}$, $y_{CG}$, $z_{CG}$ & Scaling \(\mathcal{U}(0.70, 1.30)\) \\ \hline
$W$ & Scaling \(\mathcal{U}(0.70, 1.05)\) \\ \hline
\end{tabular}
\begin{tablenotes}[para, flushleft]
\footnotesize
Modification ranges for the physical parameters—stability and control derivatives, center of gravity position, and weight—in the environment during training. Values are periodically changed throughout episodes by sampling from a uniform distribution, with limits for each parameter chosen based on domain knowledge.
\end{tablenotes}
\end{threeparttable}
\end{center}
\end{table}

\subsection{Knowledge Distillation Via Decision Transformer}

\begin{figure*}[thpb]
  \centering
  \includegraphics[scale=0.30]{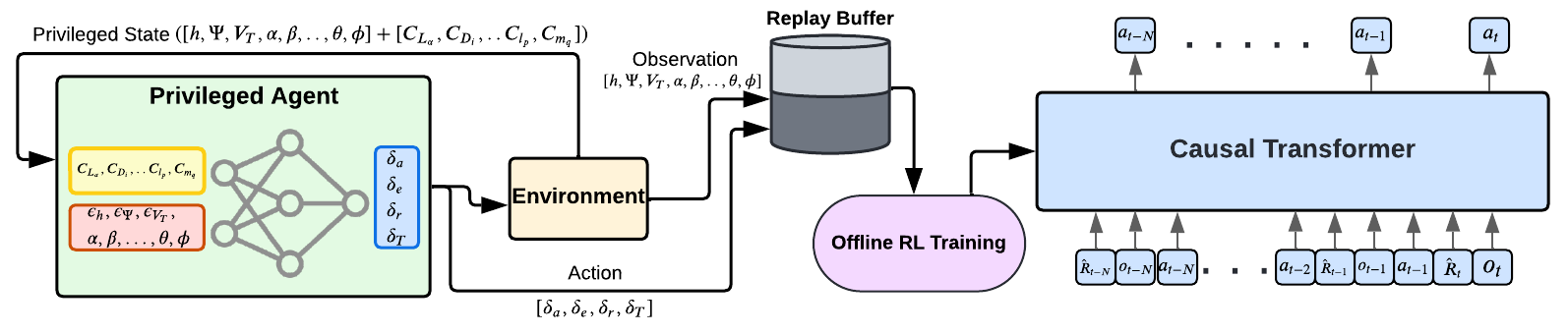}
  \caption{Knowledge distillation process through offline reinforcement learning using partial observations of the POMDP, derived from expert trajectories generated by the privileged agent.}
  \label{fig:policy_distillation}
  \vspace{-4mm}
\end{figure*}

Building upon the trained expert privileged policy, we employ offline reinforcement learning to train a student policy through knowledge distillation, using only the partial observations of the environment ($o \in O$).

As shown in Fig. \ref{fig:policy_distillation}, the privileged agent is used to collect a dataset of multiple expert trajectories, encompassing random variations in physical parameters and flight conditions. From these trajectories, we retain only the partial observations ($o_t$), taken actions ($a_t$), and rewards ($r_t$). This dataset, consisting of expert trajectories of the MDP reduced to a POMDP of partial observations, is used to train a Decision Transformer (DT) \cite{chen2021decision} through offline RL.

Decision Transformers are powerful models capable of leveraging in-context learning to adapt behavior without updating weights. Thus, the DT can learn to take optimal actions across different system dynamics using only a history of partial observations and actions. Ultimately, a DT trained on multiple expert trajectories can generalize to unseen failure scenarios and adapt its behavior to changes in the UAV dynamics using only the history of past observations to identify variations.

The final model is a causal decoder-only transformer of size $\sim$0.8M, comprising 4 heads and 3 layers, with an embedding dimension of 128 and a maximum sequence length of 60 steps (6 s). Training is performed with a batch size of 256, a learning rate of 1e-04, and a weight decay of 1e-04. The model was trained on a dataset of 2 million timesteps, encompassing 200,000 trajectories.

\subsection{Neural Lyapunov Function Learning}

To encourage stable behavior of the control policy, we incorporate a Neural Lyapunov function $V_\theta(s)$ into our training process, following ideas developed at Refs. \cite{zhou2022neural, boffi2021learning, deka2023supervised}. Inspired by \cite{deka2023supervised}, we supervise the Lyapunov function using Koopman eigenfunctions computed on the fly from trajectory data. 

Taking advantage of the trajectories dataset generated by the privileged agent, and considering as the Lyapunov state $s \in [\epsilon_{h}, \epsilon_{\Psi}, \epsilon_{V_T}, \phi, p, q, r]$, we first define an observable $f(s)$, which we choose as the L2 norm of the state:
\begin{equation}
    f(s) = \| s \|_2.
\end{equation}

The Koopman eigenfunction associated with the initial state $s_0$ is then approximated using a Laplace average along the trajectory:
\begin{equation}
    \phi(s_0) \approx \frac{\sum_{t=0}^{T-1} e^{-\lambda t}\, f(s_t)}{\sum_{t=0}^{T-1} e^{-\lambda t}},
\end{equation}
where $\lambda > 0$ is a decay parameter and $T$ is the sequence length limited to 60 steps. This weighted average provides a measure that decays along trajectories, capturing the contraction properties required of a Lyapunov function.

The Neural Lyapunov function, parameterized by a multi-layer perceptron (MLP), is trained to approximate the Koopman eigenfunction values. The MLP consists of two hidden layers with 256 neurons each, followed by ReLU activations, and an output layer with a squared activation function. The regression loss is defined as the mean squared error (MSE):
\begin{equation}
    \mathcal{L}_{\text{MSE}} = \frac{1}{N} \sum_{i=1}^{N} \left(V_\theta(s_0^{(i)}) - \phi(s_0^{(i)})\right)^2,
\end{equation}
where $N$ is the number of trajectories in the batch.

In addition to the MSE loss, we incorporate auxiliary loss terms to enforce the Lyapunov properties:
\begin{itemize}
    \item \textbf{Positive Definiteness:} We enforce $V(s) > 0$ for all $s \neq 0$ and $V(0)=0$.
    \item \textbf{Decrease Condition:} To promote stability, we penalize cases where the Lyapunov function does not decrease along the trajectory. For instance, using a hinge loss:
    \begin{equation}
        \mathcal{L}_{\text{dec}} = \frac{1}{N} \sum_{i=1}^{N} \max\{0, V_\theta(s_{t+1}^{(i)}) - V_\theta(s_t^{(i)}) + \epsilon\},
    \end{equation}
    where $\epsilon > 0$ is a margin.
\end{itemize}

The overall Lyapunov loss is then defined as a weighted combination of these terms:
\begin{equation}
    \mathcal{L}_{\text{Lyapunov}} = \mathcal{L}_{\text{MSE}} + \alpha\, \mathcal{L}_{\text{dec}},
\end{equation}
with $\alpha$ being a weighting factor chosen empirically. The learned Lyapunov function can be used to study the stability guarantees of a trained controller.

Finally, to encourage stability in the decision-making process, the learned neural Lyapunov function is incorporated into the loss of the Decision Transformer. The final loss for the Decision Transformer becomes:
\begin{equation}
    \mathcal{L}_{\text{total}} = \mathcal{L}_{\text{DT}} + \gamma\, \mathcal{L}_{\text{Lyapunov}},
\end{equation}
where $\mathcal{L}_{\text{DT}}$ denotes the original Decision Transformer loss and $\gamma$ is a hyperparameter that balances the contribution of the Lyapunov loss. By minimizing $\mathcal{L}_{\text{total}}$, the Decision Transformer is guided to select actions that not only maximize reward but also lead the system towards more stable states as indicated by a decreasing $V_\theta(s)$.

\section{Experiments}

In this section we present several experiments to validate the effectiveness of our proposed method. We first provide a description of the testing cases designed to validate our method in specific failure scenarios rather than randomized variations, as well as the implementation details of the experiments. Secondly, we demonstrate our model's capability through a comparative study across different cases.

\subsection{Testing Cases and Implementation Details}

We evaluate our method across several failure and damage scenarios for the UAV. Unlike the random and independent parameter modifications used during training, testing cases involve modifications to aerodynamic parameters and control surfaces based on the changes that a specific failure produces in the UAV's dynamics, creating unseen situations for the agent. The defined scenarios, inspired by Refs. \cite{dally2022soft, gavra2024evolutionary}, are:

\begin{itemize}
    \item \textit{Jammed Rudder}: Rudder stuck at a deflection of 15°.
    \item \textit{Broken Aileron}: Aileron effectiveness decreased by 50\% with an increase in the parasitic roll moment due to asymmetric force.
    \item \textit{Saturated Elevator}: Elevator deflection limited to $\pm$5\% of the maximum range.
    \item \textit{Deployed Landing Gear}: Sudden deployment of landing gear, resulting in increased drag given by $3C_{D_\text{gear}}$, which was unseen during training.
    \item \textit{Shifted Center of Gravity}: UAV center of gravity shifts along the $x$, $y$, and $z$ axes by +20\%, +40\%, and -30\%, respectively.
      \item \textit{Damaged Horizontal Tail}: Partial tail damage causing reduced lift ($0.5C_{L_{\delta_e}}$), increased drag ($1.2C_{D_{\delta_e}}$, $1.2C_{D_0}$), and affecting pitch moment ($0.5C_{m_{\delta_e}}$, $C_{m_0}$, $0.5C_{m_q}$, $0.5C_{m_\alpha}$, $0.5C_{m_{\dot{\alpha}}}$).
    \item \textit{Damaged Semi-Wing}: Damaged wing producing a reduction in lift effectiveness ($0.7C_{L_\alpha}$, $0.7C_{L_q}$, $0.7C_{L_{\dot{\alpha}}}$), increased drag ($1.2C_{D_0}$, $1.2C_{D_i}$), and affected side force ($1.5C_{Y_\beta}$, $1.5C_{Y_p}$, $1.5C_{Y_r}$). Roll moment altered ($C_{l_0}$, $-1.0C_{l_\beta}$, $-1.0C_{l_r}$, $0.5C_{l_{\delta_a}}$), and yaw moment modified ($C_{n_0}$, $0.8C_{n_r}$, $1.5C_{n_\beta}$, $1.5C_{n_{\delta_a}}$, $1.2C_{n_\alpha}$).
\end{itemize}

\vspace{0.1cm}

During validation, multiple episodes are run in which the UAV's dynamics are changed from the nominal condition to the modifications defined for each failure scenario at a specified timestep. Set-point changes for ($h, \Psi, V_T$) also occur at specified timesteps.

Experiments and training are conducted using JSBSim, a high-fidelity 6-DoF flight dynamics model. The UAV physics model used operates within flight conditions of $h \in [4000, 24000]$ ft and $V_T \in [260, 360]$ ft/s. Episodes have a maximum length of 1000 steps. To account for real-world uncertainties, we incorporate external perturbations and sensor noise during both training and testing. Wind gusts are applied following a Gaussian distribution \textit{N(0, 2 ft/s)}, introducing stochastic disturbances. Additionally, observation noise --modeled by a Gaussian distribution \textit{N(0, 0.02x)}-- is injected into sensor measurements to simulate real-world inaccuracies, ensuring the robustness of both the teacher and student models. These measures help bridge the simulation to real (sim-to-real) gap, improving generalization to real UAV conditions and paving the way for future real-world validation.

\subsection{Comparison Study}

\begin{table*}[ht]
\caption{Comparison Study Results}
\label{tab:comparison_study_results}
\begin{center}
\begin{threeparttable}
\begin{tabular}{ccccccc}
\hline
\textbf{Scenarios} & & \textbf{DT (Ours)} & \textbf{RL+DR} & \textbf{RL} & \textbf{FCS} & \textbf{FCS+RL} \\ \hline
\multirow{2}{*}{\textbf{Nominal}} & \textbf{$\mu \pm \sigma$} & \textbf{-124.61 ± 25.25} & -232.59 ± 90.27 & -205.53 ± 46.71 & -257.30 ± 34.15 & -259.80 ± 36.42 \\
 & \textbf{Crash \%} & 0.00\% & 0.00\% & 0.00\% & 0.00\% & 0.00\% \\ \hline
\multirow{2}{*}{\textbf{Jammed Rudder}} & \textbf{$\mu \pm \sigma$} & \textbf{-398.49 ± 25.69} & -409.40 ± 55.31 & -412.40 ± 79.35 & -585.60 ± 393.28 & -543.04 ± 390.62 \\
 & \textbf{Crash \%} & 0.00\% & 0.00\% & 0.00\% & 20.00\% & 32.00\% \\ \hline
\multirow{2}{*}{\textbf{Broken Aileron}} & \textbf{$\mu \pm \sigma$} & \textbf{-151.94 ± 26.67} & -398.75 ± 70.36 & -409.06 ± 24.57 & -485.19 ± 211.83 & -430.50 ± 257.86 \\
 & \textbf{Crash \%} & 0.00\% & 0.00\% & 0.00\% & 4.00\% & 16.00\% \\ \hline
\multirow{2}{*}{\textbf{Saturated Elevator}} & \textbf{$\mu \pm \sigma$} & \textbf{-220.11 ± 74.98} & -301.69 ± 54.27 & -301.84 ± 69.89 & -545.53 ± 35.78 & -443.64 ± 31.37 \\
 & \textbf{Crash \%} & 0.00\% & 0.00\% & 0.00\% & 0.00\% & 0.00\% \\ \hline
\multirow{2}{*}{\textbf{Deployed Landing Gear}} & \textbf{$\mu \pm \sigma$} & \textbf{-216.02 ± 73.54} & -307.10 ± 77.71 & -270.45 ± 27.50 & -552.65 ± 306.27 & -382.85 ± 92.94 \\
 & \textbf{Crash \%} & 0.00\% & 0.00\% & 0.00\% & 12.00\% & 8.00\% \\ \hline
\multirow{2}{*}{\textbf{Shifted CG}} & \textbf{$\mu \pm \sigma$} & \textbf{-142.78 ± 19.45} & -349.35 ± 94.51 & -272.70 ± 51.39 & -733.69 ± 420.98 & -414.10 ± 103.58 \\
 & \textbf{Crash \%} & 0.00\% & 0.00\% & 0.00\% & 28.00\% & 20.00\% \\ \hline
\multirow{2}{*}{\textbf{Damaged Tail}} & \textbf{$\mu \pm \sigma$} & \textbf{-202.10 ± 40.86} & -430.47 ± 46.02 & -349.36 ± 28.81 & -721.62 ± 436.57 & -402.62 ± 232.80 \\
 & \textbf{Crash \%} & 0.00\% & 0.00\% & 0.00\% & 28.00\% & 8.00\% \\ \hline
\multirow{2}{*}{\textbf{Damaged Wing}} & \textbf{$\mu \pm \sigma$} & \textbf{-318.68 ± 20.46} & -379.25 ± 91.68 & -1394.93 ± 155.40 & -1167.78 ± 338.98 & -762.48 ± 444.57 \\
 & \textbf{Crash \%} & 4.00\% & 9.00\% & 100.00\% & 92.00\% & 60.00\% \\ \hline
\end{tabular}
\begin{tablenotes}[para, flushleft]
\footnotesize
Comparison of performance results across different algorithms. DT refers to our method. RL+DR denotes the DreamerV3 agent trained with Domain Randomization (DR) without privileged environment information. RL refers to the DreamerV3 agent assuming an MDP formulation. FCS represents the baseline industry-like designed FCS. FCS+RL refers to the FCS with added Gain-Scheduler (GS) trained using RL. Performance is measured using the mean and standard deviation of episode returns and the crash percentage across 100 episodes for each failure scenario.
\end{tablenotes}
\end{threeparttable}
\end{center}
\end{table*}

We evaluate our proposed method against several state-of-the-art flight control systems (FCS) to showcase the advantages of our approach in fault-tolerant flight control.

\begin{itemize}
    \item \textit{Privileged Policy}: The teacher policy trained with privileged information of the environment.
    \item \textit{RL Policy}: RL-trained policy without environmental changes, assuming an MDP. This policy is tested in the POMDP setting and trained using DreamerV3.
    \item \textit{RL Policy + DR}: RL policy trained with environment variations through domain randomization, considering the setup as a POMDP.
    \item \textit{Industry FCS}: A baseline FCS following common industry practices, with separate control for lateral-directional and longitudinal motions using nested PID controllers.
    \item \textit{Industry FCS + RL + DR}: A RL policy trained to select the PID gains of the FCS at each step, following a Gain-Scheduling (GS) approach for fault-tolerant control. The policy is trained in the POMDP setting with domain randomization.
\end{itemize}

\begin{figure*}[thpb]
  \centering
  \begin{subfigure}{0.48\textwidth}
    \centering
    \includegraphics[width=\linewidth]{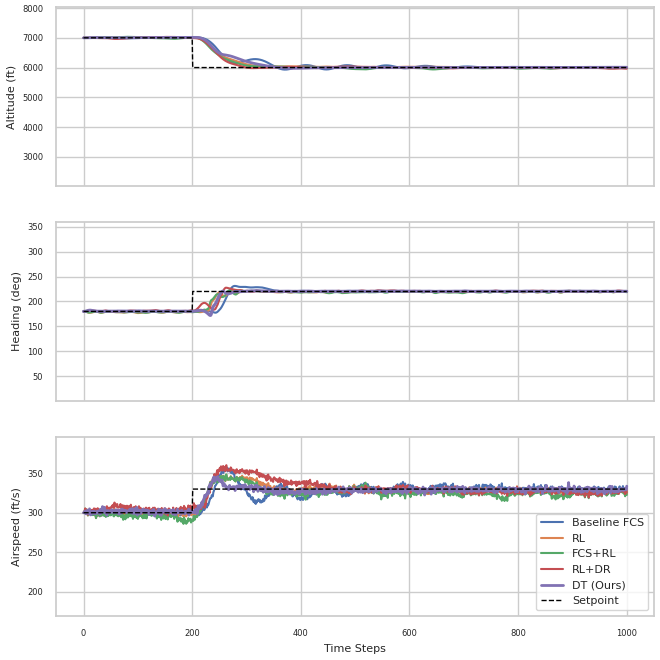}
    \caption{}
    \label{fig:traj_nominal}
  \end{subfigure}
  \begin{subfigure}{0.48\textwidth}
    \centering
    \includegraphics[width=\linewidth]{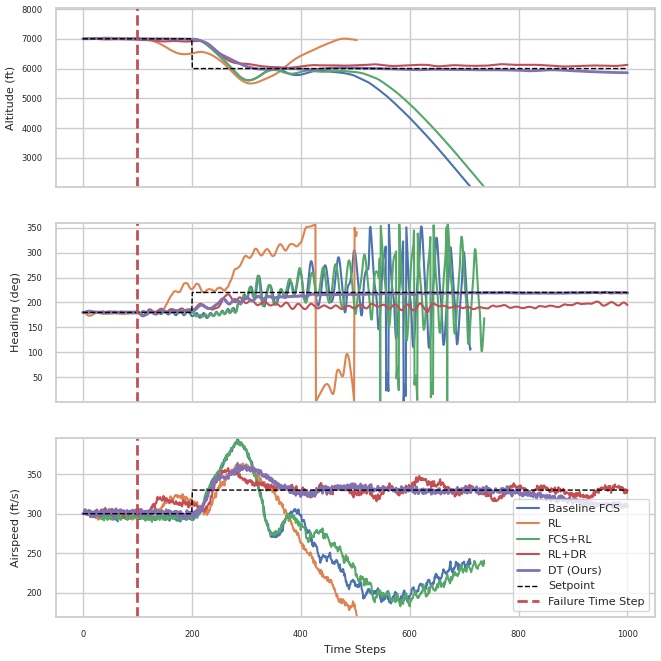}
    \caption{}
    \label{fig:traj_wing}
  \end{subfigure} 
  \caption{Sample trajectories comparing our DT method against the baseline FCS, RL trained with DR (RL+DR), base RL agent (RL), and trained GS system (FCS+RL) in the nominal and damaged wing scenarios for tracking reference values of altitude ($h$), heading ($\Psi$), and airspeed ($V_T$). (a) Nominal scenario. (b) Damaged wing scenario. The dashed black line represents the setpoint for each value, and the vertical dashed red line marks the timestep at which the failure occurs.}
  \label{fig:trajectories}
\end{figure*}

\begin{figure}[thpb]
  \centering
  \includegraphics[scale=0.42]{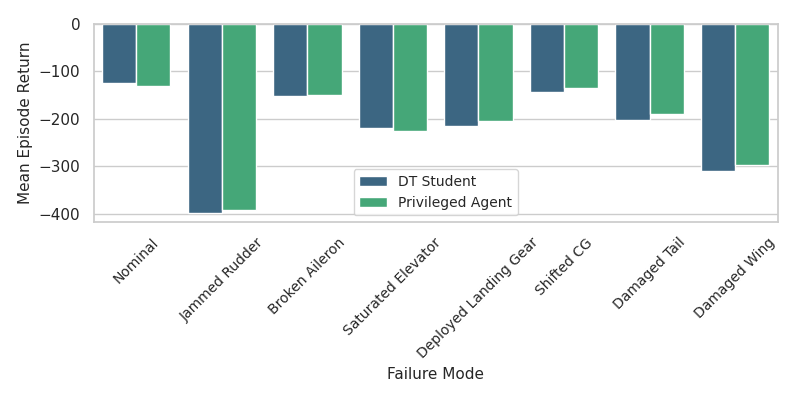}
  \caption{Comparison between the privileged agent teacher and DT student policies across each failure scenario, using mean episode return as the performance metric.}
  \label{fig:comp_dt_failaware}
\end{figure}

To assess the effectiveness of the DT under POMDP conditions, we compare the results of the DT student agent and the privileged teacher agent. Both policies are tested across all scenarios in 100 episodes each, with flight conditions and random seeds standardized for fair comparison. Fig. \ref{fig:comp_dt_failaware} shows the results, using the mean episode return as the performance metric, where the theoretical maximum is zero. Results demonstrate similar performance between the agents, showing the DT's capability to learn behaviors from the expert dataset under the POMDP setting and to adapt to dynamic variations in context. The student model's ability to match or outperform the teacher despite limited observations arises from the nature of offline reinforcement learning (RL). Rather than directly imitating the teacher, the student extracts essential decision patterns from expert trajectories, prioritizing key contextual cues for robust adaptation. This aligns with the findings in Ref. \cite{chen2021decision}, where DTs demonstrate the ability to outperform their training data by using return-conditioning to generalize more effectively under uncertainty. Despite privileged information, the teacher may suffer from unstable training, whereas the student, trained on diverse replay data, learns a more adaptive strategy for handling unforeseen failures. Moreover, incorporating the Lyapunov-based loss term into the offline RL training further encourages the DT student to select actions that lead to more stable states, thereby improving overall return.

\begin{table}[ht]
\caption{Student Model Architectures}
\label{tab:student_architectures}
\begin{center}
\begin{threeparttable}
\begin{tabular}{ccccccc}
\hline
\textbf{Scenarios} & & \textbf{DT (Ours)} & \textbf{LSTM} & \textbf{TCN} \\ \hline
\multirow{3}{*}{\textbf{Nominal}} & \textbf{$e_{h}$} & \textbf{63.3 ± 55.4} & 102.3 ± 34.2 & 124.1 ± 41.7 \\
& \textbf{$e_{\Psi}$} & \textbf{2.4 ± 2.3} & 3.5 ± 2.2 & 3.9 ± 2.9 \\
& \textbf{$e_{V_T}$} & \textbf{4.6 ± 3.4} & 6.5 ± 3.7 & 9.1 ± 4.9 \\
\hline
\multirow{3}{*}{\textbf{D. Wing}} & \textbf{$e_{h}$} & \textbf{176.9 ± 90.6} & 1265.3 ± 1104.1 & 1823.6 ± 1331.9\\
& \textbf{$e_{\Psi}$} & \textbf{12.8 ± 6.6} & 14.7 ± 5.3 & 16.4 ± 5.8 \\
& \textbf{$e_{V_T}$} & \textbf{9.4 ± 5.9} & 54.9 ± 40.5 & 69.3 ± 43.1 \\
\hline
\end{tabular}
\begin{tablenotes}[para, flushleft]
\footnotesize
Tracking error performance comparison (in mean and std) between DT (Ours) and LSTM and TCN architectures. Errors are measured as L2 distance. $e_h$ in ft, $e_{\Psi}$ in degrees and $e_{V_T}$ in ft/s.
\end{tablenotes}
\end{threeparttable}
\end{center}
\end{table}

While transformers require higher computational resources than traditional sequence models, their parallelization may enable efficient inference, making them feasible for real-time UAV control in adequate hardware. Our model remains lightweight, ensuring real-time feasibility, as has been demonstrated in similar-size models in Ref. \cite{Burrello2021Microcontroller, Jung2024TinyTransformers, Yang2024MCUBERT, Liang2023MCUFormer}. Moreover, a trade-off between computational overhead and performance should be made. As shown in Table \ref{tab:student_architectures}, our DT achieves significantly lower tracking errors than LSTM and Temporal Convolutional Network (TCN), especially in failure scenarios. These results validate the superiority of transformer-based control in both accuracy and robustness.

\begin{figure}[htbp]
  \centering
  \begin{subfigure}{0.4\textwidth}
    \centering
    \includegraphics[width=\linewidth]{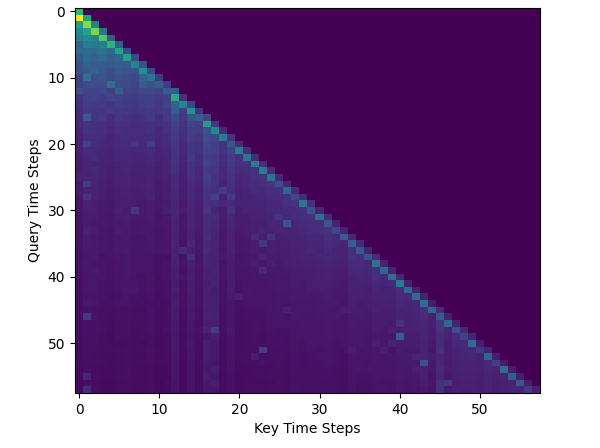}
    \caption{}
    \label{fig:att_map}
  \end{subfigure}
  \begin{subfigure}{0.4\textwidth}
    \centering
    \includegraphics[width=\linewidth]{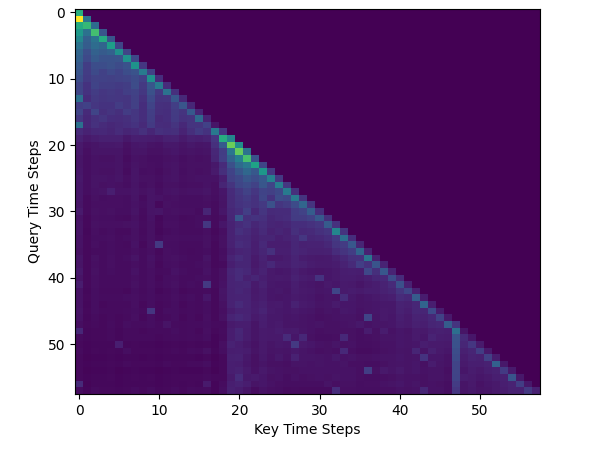}
    \caption{}
    \label{fig:att_map_failure}
  \end{subfigure}
  \caption{Attention maps of the Decision Transformer (DT) before and after a failure occurs. Attention values are averaged across heads, layers, and $(o_t, a_t, r_t)$ tuples over the 60-step context window. (a) Nominal case before failure, showing evenly distributed attention across past time steps. (b) Post-failure case (failure at timestep 100, current time at 140)}
  \label{fig:att_maps}
  
\end{figure}

One key concern with transformer-based control systems in safety-critical applications is their black-box nature. To improve interpretability, we analyze the attention mechanism of the Decision Transformer (DT) to understand how it identifies failures and adapts its control policy. Figure \ref{fig:att_maps} shows the attention patterns before and after a failure occurs, over the 60-step context window. In the nominal case (Figure \ref{fig:att_map}), attention is distributed evenly across past states. However, after a failure at timestep 100 (where the current timestep is 140), the DT adapts by focusing more on the time steps following the failure (Figure \ref{fig:att_map_failure}), indicating a shift in control strategy based on recent dynamics. This adaptation mechanism enhances robustness by dynamically weighting relevant information in response to sudden changes in UAV behavior.

A full comparison study, as shown in Table \ref{tab:comparison_study_results}, assesses the DT's performance against other algorithms, including state-of-the-art online RL agents and industry-standard FCS commonly used in commercial UAVs. Performance is measured by the mean and standard deviation of episode returns, as well as crash percentage due to total control loss. Fatal failures terminate an episode, occurring under extreme values of angular rates or accelerations, or when the UAV breaches lower altitude limits. As before, 100 episodes are run for each failure scenario. Results in Table \ref{tab:comparison_study_results} show that the DT policy outperforms others in all failure and nominal scenarios, effectively tracking set-points and maintaining control even under severe damage and extreme conditions.

Promising results are also obtained from the RL agent with domain randomization, showing high tracking performance and achieving zero crashes across failure scenarios. These results highlight the potential of model-based RL algorithms with added memory, such as DreamerV3. However, the DT agent’s superior performance, despite its significantly smaller parameter size ($\sim$0.8M) compared to DreamerV3 ($\sim$26M), underscores the potential of transformer-based models for real-time dynamics adaptation through in-context learning.

\begin{figure}[thpb]
  \centering
  \includegraphics[scale=0.45]{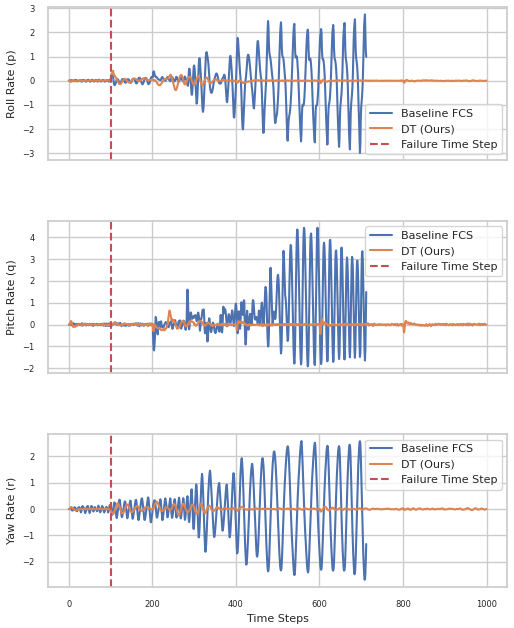}
  \caption{Angular rates ($p$, $q$, $r$) following wing damage, comparing DT agent and baseline FCS responses. The dashed red line marks the time step at which failure occurs.}
  \label{fig:traj_rates}
\end{figure}

Currently used FCSs, as described, are difficult to replace due to their simplicity, reliability, modularity, and computational efficiency, all while avoiding the need for an accurate UAV model and handling uncertainties. However, they lack adaptability to significant dynamic changes. Adding a Gain Scheduling (GS) algorithm to the FCS could enhance system adaptability under failure conditions by modifying gains in real time; however, GS faces challenges related to computational demands and nonlinear dynamics. Even state-of-the-art RL algorithms trained for fault tolerance struggle to achieve high performance across scenarios, as shown in Table \ref{tab:comparison_study_results}.

Our method takes advantage of the in-context learning capabilities of transformers to adapt policy behavior during flight without changing weights \cite{radosavovic2024real}, rather than modifying controller gains. Fig. \ref{fig:trajectories} presents sample trajectories of our method versus the other algorithms in the comparison study for tracking $h$, $\Psi$, and $V_T$. In Fig. \ref{fig:traj_nominal}, the nominal UAV dynamics are shown, where all designed controllers track reference values with similarly high performance, with some noise in the measured airspeed due to light wind gusts.

Fig. \ref{fig:traj_wing} shows the trajectory for the damaged wing scenario, the most challenging due to the significant dynamic changes. In this sample, the failure occurs at timestep 100 (indicated by the vertical dashed red line), and the setpoint change requirement occurs at timestep 200. While the Baseline FCS, the base RL agent, and the FCS with added GS (FCS+RL) diverge, losing control and resulting in a crash and episode termination (triggered by either extreme accelerations or entering a tailspin), the RL agent trained with DR (RL+DR) and our DT method regain control and track setpoints effectively despite substantial damage. Among these, the DT method achieves the highest tracking accuracy for all three reference values showing superior performance.

Fig. \ref{fig:traj_rates} illustrates angular rates ($p$, $q$, $r$) for our DT method and the baseline FCS in the damaged wing scenario. The vertical red line marks the onset of failure, highlighting the abrupt disturbance caused in the roll rate ($p$). After this point, the controller must regain control of the UAV to re-stabilize it to the previous flight condition. In this case, our DT agent successfully stabilizes the UAV, converging to the prior steady flight state despite the damaged aerodynamics and control surfaces. In contrast, the baseline industry-designed FCS begins to diverge due to accumulating errors over time, eventually leading to total control loss and stall.

\subsection{Stability Analysis}

Under partial observability and lacking an explicit model-based design, conventional methods for guaranteeing stability may be theoretically infeasible. However, following the findings in Refs.~\cite{liu2024tool, zhou2022neural, boffi2021learning, han2020actor, deka2023supervised}, we can use the learned Neural Lyapunov function to derive partial guarantees of stability and safety. Because the learned function satisfies the properties of a Lyapunov function, a controller that drives the system towards stability must exhibit a decreasing Lyapunov function, \(V_\theta(s)\), throughout the episode, eventually reaching zero if the system is fully stabilized at 
\[
s \in [\epsilon_{h}, \epsilon_{\Psi}, \epsilon_{V_T}, \phi, p, q, r] = \hat{0}.
\]

To further examine the impact of including the Lyapunov loss term \(\mathcal{L}_{\text{Lyapunov}}\) in Decision Transformer (DT) training, we evaluate the Lyapunov-based stability of both a baseline DT and the Lyapunov-enhanced DT. We conduct 100 experiments for each model under both nominal and failure scenarios (combining different types of failures).

\begin{figure}[htbp]
  \centering
  \begin{subfigure}{0.49\textwidth}
    \centering
    \includegraphics[width=\linewidth]{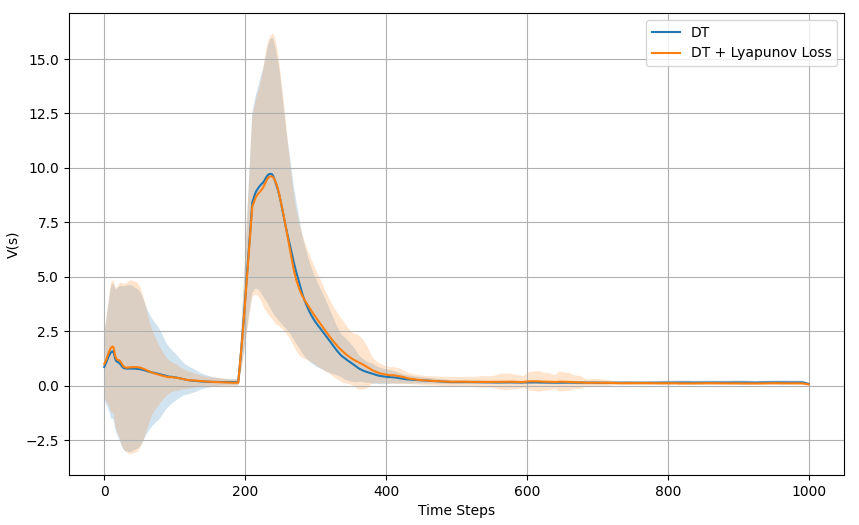}
    \caption{}
    \label{fig:stability_analysis_nominal}
  \end{subfigure}
  \hfill
  \begin{subfigure}{0.49\textwidth}
    \centering
    \includegraphics[width=\linewidth]{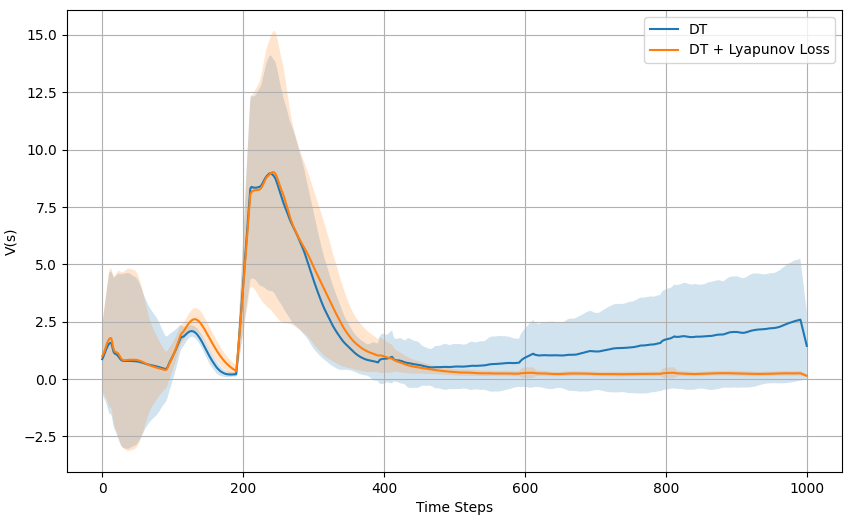}
    \caption{}
    \label{fig:stability_analysis_failure}
  \end{subfigure}
  \caption{Mean and standard deviation of the Neural Lyapunov function, \(V_\theta(s)\), over time for the baseline DT and the Lyapunov-enhanced DT. (a)~Nominal conditions, showing both models converging to minimal energy; (b)~Failure scenario, illustrating how the Lyapunov-enhanced DT maintains lower energy, indicating greater stability.}
  \label{fig:stability_analysis}
\end{figure}

Fig.~\ref{fig:stability_analysis} presents the mean and standard deviation of the Neural Lyapunov function, \(V_\theta(s)\), over the timesteps in an episode—providing a measure of the “energy” in the system. If the Lyapunov energy tends to zero, it indicates that the system is converging to a stable equilibrium state. In Fig.~\ref{fig:stability_analysis_nominal}, we observe that under nominal conditions, both DT models (trained with and without Lyapunov) display asymptotic stability, converging to minimal energy. The spike at timestep 200 arises from an intentional change in the reference values for altitude, heading, and airspeed (\(h, \Psi, V_T\)); after this disturbance, both models settle into a stable state.

Fig.~\ref{fig:stability_analysis_failure} shows the failure scenario. A failure occurs at timestep 100, causing a surge in energy that both models stabilize quickly. However, once new reference values are introduced at timestep 200, the Lyapunov-enhanced DT converges to near-zero energy, while the baseline DT’s energy gradually increases until the episode ends. This behavior indicates that incorporating Lyapunov-based ideas into DT training can promote stability at inference.

\subsection{Discussion}

While several nonlinear fault-tolerant control methods have been proposed—such as those based on fuzzy logic \cite{yao2019fault} and sliding mode control \cite{ezzara2024sliding}—their adoption in real-world applications remains limited. One of the main reasons is that these approaches often require complex, computationally intensive routines for continuously monitoring system dynamics and adjusting control laws online. This computational overhead is a critical drawback for embedded platforms, where processing power and memory are tightly constrained.

In contrast, traditional cascade PID controllers, which are employed in widely used autopilot systems such as PX4 and Ardupilot, offer a much simpler and computationally efficient solution for low-level attitude control. Their hierarchical structure minimizes the need for online computations and parameter re-tuning, making them particularly well suited for microcontroller-based implementations. However, the simplicity of cascade PIDs also limits their performance when the UAV dynamics deviate significantly from the nominal model.

Our approach addresses these computational constraints by shifting the heavy computational burden to the offline training phase. The resulting Decision Transformer (DT) model, with only around 0.8M parameters, is designed for efficient real-time inference. Notably, recent studies have demonstrated that transformer models of similar size can be successfully deployed on microcontrollers—such as ARM Cortex-M devices—through aggressive quantization techniques that reduce both memory footprint and computational load \cite{Burrello2021Microcontroller, Jung2024TinyTransformers, Yang2024MCUBERT, Liang2023MCUFormer}. With quantization (e.g., converting weights and activations to 8-bit representations), the DT model’s inference can be optimized to meet the real-time constraints imposed by onboard flight controllers.

By eliminating the need for online re-parameterization, weight adaptation, or explicit fault identification during operation, our DT model circumvents the latency and computational complexity inherent in many nonlinear fault-tolerant control methods. This makes it a promising alternative for robust UAV control, effectively bridging the gap between advanced learning-based strategies and the stringent hardware limitations of embedded systems.


\section{Conclusion}

In this paper, we introduced a transformer-based fault-tolerant flight control framework for fixed-wing UAVs that bypasses classical inner-loop controllers, drastically reducing the complexity of real-time adaptation to system failures. By coupling teacher-student knowledge distillation with a Decision Transformer trained in an offline RL setting, we demonstrated that partial observability can be effectively mitigated: the student learns robust policies solely from past state histories without explicit failure detection or online weight adaptation.

Our experiments reveal that the proposed controller achieves superior performance in both nominal and failure scenarios. In comparison with industry-standard PID-based FCS and other state-of-the-art RL agents, the transformer policy exhibited enhanced tracking accuracy and reduced crash rates, even under severe aerodynamic disruptions such as wing and tail damage. Through its attention mechanism, the Decision Transformer reallocates focus to post-failure observations, thereby adapting its control actions in context. This design also circumvents the need for real-time reparametrization or scheduled gain updates, offering a streamlined alternative to conventional fault-tolerant methods.

By incorporating Lyapunov-based constraints into the offline training process, the Decision Transformer is encouraged to explore actions that promote overall system stability, further improving safety in extreme conditions. The resulting lightweight model—comprising fewer than one million parameters—can be deployed efficiently on embedded processors, especially when combined with quantization. Our results thus highlight the potential of transformer architectures to unify real-time adaptation, fault tolerance, and computational feasibility in UAV control.

Future work will focus on hardware-in-the-loop and real-flight experiments to validate these promising simulation results under practical constraints such as limited onboard computation, sensor noise, and stringent flight-envelope requirements. Overall, this work serves as a step toward reliable, adaptive UAV operations that blend the robustness of classical control with the flexibility of modern deep learning methods, paving the way for safer and more versatile aerial robotic systems.

\section{Acknowledgments}

We thank Miguel Gómez-López and Rodney Rodríguez Robles for useful discussions on control theory. 

S.L.C. acknowledges the grant PID2023-147790OB-I00 funded by MCIU/AEI/10.13039 /501100011033 /FEDER, UE. S.L.C. also acknowledges the MODELAIR project that have received funding from the European Union’s Horizon Europe research and innovation programme under the Marie Sklodowska-Curie grant agreement No. 101072559. The results of this publication reflect only the author's view and do not necessarily reflect those of the European Union. The European Union can not be held responsible for them. 

R.V. acknowledges financial support from ERC grant no.2021-CoG-101043998, DEEPCONTROL. Views and opinions expressed are however those of the author(s) only and do not necessarily reflect those of the European Union or the European Research Council. Neither the European Union nor the granting authority can be held responsible for them.

\addtolength{\textheight}{-13.5cm}   


\bibliographystyle{IEEEtran}
\bibliography{IEEEabrv,bibliography}


 





\end{document}